# MambaFormer: Token-Level Guided Routing Mixture-of-Experts for Accurate and Efficient Clinical Assistance

Hamad Khan[1], Saddam Hussain Khan[1]*

[1]Artificial Intelligence Lab, Department of Computer Systems Engineering, University of Engineering and Applied Sciences (UEAS), Swat 19060, Pakistan

**Email:** hengrshkhan822@gmail.com

## ABSTRACT

The deployment of large language models (LLMs) in real-world clinical applications is constrained by the fundamental trade-off between computational cost and the efficiency of linear-time models. To address this, we propose an LLM-based MambaFormer hybrid Mixture-of-Experts (MoE) framework for efficient medical question-answering (QA) and clinical assistance. The MambaFormer employs a lightweight gating mechanism that performs token-level dynamic routing to a customized Transformer expert ($E_{T5}$) for short, complex queries or to a State Space Model expert ($E_{Mamba}$) for long, high-throughput sequences. The customized $E_{Mamba}$ and $E_{T5}$ models are adapted to input sequence dimensionality, embedding structure, sequence length, and target-specific output heads, and fine-tuned through transfer learning on a new, custom-designed DentalQA dataset. Moreover, intelligent routing decisions are driven by the contextual complexity of token embeddings, normalized sequence length, and domain-aware features, thereby enforcing a Pareto-optimal trade-off between inference latency and prediction accuracy. Furthermore, a novel utility-guided multi-objective loss jointly optimizes decisions, router parameters, routing behavior, expert utilization, and computational cost by adaptively regulating token-level expert activation. Finally, the proposed MambaFormer is cross-validated (holdout) for medical QA on the new, custom-designed DentalQA and PubMedQA datasets and compared with state-of-the-art techniques. The proposed MambaFormer outperforms (BERTScore = 0.9180) with ultra-low latency (0.077 s), delivering a 24.4× speedup over T5-Large and establishing a scalable solution for resource-constrained clinical deployment.

# 1. Introduction

Natural language processing has shown significant advancements due to the rise of large language models, which improve performance through large-scale pretraining [1]. NLP effectiveness in complex reasoning remains constrained by limited grounding and high computational cost [2]. This also hampers continual knowledge updates and raises sustainability concerns [3], [4]. NLP plays an increasingly important role in clinical applications such as risk assessment, therapeutic support, and disease prediction, despite these challenges [5], [6]. However, NLP faces particular difficulties in the therapeutic sector, such as the intricacy of medical language, as well as in computational resources.

Various preeminent language models have been introduced. Traditional models like Transformer [7] were introduced, but the transformer uses a self-attention mechanism using a formula $O(N^2)$, where N is the sequence length. Though this model is satisfactory in terms of accuracy, it lacks in terms of memory and speed, resulting in high computational resources. This was tested by detailed clinical reports [8] and similarly hinders its deployment in other data-heavy domains like medical image segmentation [9]. In real-time applications and hospitals, time is a very important factor, and due to the above situation, for example, a system may take 1.8 seconds to run 1000 words of clinical information, which is too slow, and for hospitals and clinics, quick processing time is important. There is a need for a system with a quick processing time, and it has to keep the privacy of all patients' information locally and globally. Other transformer models like BioBERT [10], and ClinicalBERT [5] also succeeded on biomedical tasks with a BERT score of 90% but this results in high costs.

In addition to diminishing this constraint, models like LongFormer [11] and BigBird [12] used sparse attention, which reduces intricacy to $O(N)$ using attention patterns. Kernel-based methods, such as Performer [13], were used, which leads to achieving efficiency gains better than the transformer. However, when tested on long medical notes, it shows an accuracy drop of 5 to 10%, which is not suitable for use in a clinical environment. More recently, State Space Models (SSMs) have been recognized, particularly the Mamba architecture [14], Very recent developments (2024) have introduced advanced SSM variants like StripedHyena [15] and MambaByte [16], which further optimize selective scanning mechanisms. Concurrently, hybrid architectures such as MoE-Mamba [17] and Block-State Transformers [18] explore different fusion strategies between attention and recurrence, though they typically employ fixed or block-level routing rather than token-level adaptation. as depending on the Selective state space Model (SSM) as a better alternative due to their linear scaling and longer dependency modeling capabilities, as shown in Equation (1), where L denotes the input sequence length. Mamba analyzes lengthy sequences with great efficiency and at a reduced cost of training and inference [19]. However, it exhibits inconsistent related learning and dual-task generalization abilities, and it also struggles to capture locally important information in text. Similarly, Transformer shows favourable results in accuracy, but in short sequences of length, it demands high computational resources and also costs [7]. This creates a major conflict in clinical applications where both accuracy and speed are essential. Yet there again rises a problem for medical doctors, either to use a transformer for accuracy, but will face computational cost, or will lose performance by sticking with SSM.

The Pareto trade-off has not been fully solved and has been struggled with by any of the individual or even hybrid models. Hybrid models like Jamba and MoE [20], [21], which are an integration of Transformers and SSM, show a positive sign in this trade-off. However, even the most recent hybrid approaches continue to struggle with adaptive routing, relying on predetermined architectural patterns rather than context-aware token-level decisions. This gap persists despite advances in efficient LLMs like Mistral 7B and Phi-3 [22], [23], which demonstrate the practical viability of MoE designs but lack fine-grained routing mechanisms suitable for clinical text. However, their design is based on a fixed routing and does not contain any smart rules or penalties for the transformer or SSM to show the computational reports. Even most advanced models like Mamba-3 [24] and RetNet-MoE [25] continue to use predetermined routing patterns rather than context-aware token-level decisions, leaving a gap that our MambaFormer addresses. In addition, they have any information on which expert will be used for accuracy and speed, which overcomes all aforementioned limitations and provides a complete Model that is capable of efficiency and cost. Therefore, a smarter routing mechanism is needed to dynamically balance accuracy and efficiency. In this regard, we have introduced Proposed MambaFormer, a utility-guided dynamic routing framework for Pareto-optimal efficiency in BioNLP. Our novel approach makes the following contributions:

- Our novel hybrid Mixture-of-Experts (MoE) framework, 'MambaFormer', is the integration of customized SSM Mamba ($E_{Mamba}$) and Transformer ($E_{T5}$) that employs a new guided gating mechanism to dynamically route each incoming token for clinical assistance. Moreover, MambaFormer introduced a new multiobjective function ($L_{MambaFormer}$) that automatically optimizes the adaptive routing parameters, which systematically assign tokens to the optimal experts.
- The customized SSM Mamba ($E_{Mamba}$) and Transformer ($E_{T5}$) input layers are modified according to input token dimensionality, embedding structure, and length, and output layers are added to target-specific QA and description. Moreover, these models are fine-tuned through transfer learning on a new custom-designed DentalQA comprising 5,000 QA pairs (10,000 tokens) and the PubMedQA dataset.
- The MambaFormer systematically integrates token features with a variational sequence to either a $E_{T5}$(complex short contents, high-accuracy cases) or a highly-efficient $E_{Mamba}$ expert (long sequences, high-throughput cases) through a proposed intelligent router. Moreover, the steering-guided router learns domain-aware and sequence-length features to dynamically direct tokens to the optimal expert based on a Pareto-optimal trade-off between accuracy and latency. As a result, 96.2% of tokens are provided to the fast and efficient $E_{Mamba}$ expert, while only 3.8% of tokens go to the high accuracy $E_{T5}$ expert existing dataset diversity.

- The proposed MambaFormer framework successfully surpasses traditional Models in speed (response time) by gaining 0.077 seconds (a 24.4x speedup over traditional models) while maintaining BERTScore F1 of 0.9180 and accuracy of 96.2%.

The rest of the manuscript is organized as follows: Section 2 presents an overview of previous studies, and Section 3 outlines the methodology. Section 4 describes the experimental setup, Section 5 presents the results, and Section 6 concludes the study.

## 2. Related Work

Numerous studies have examined the intricacy of efficient biological NLP, including Transformer models and other complicated hybrid deep learning models. Current NLP systems are based on transformer scaling, which was first introduced by Scaled Multiplicative-attention Vaswani et al. [7], and made it possible to record global dependencies without repetition. Following this, BERT [26] was used, which demonstrated the effectiveness of bidirectional pre-training by achieving 90% F1 scores on the named entity recognition baseline. Text-to-text challenges were then used to represent NLP tasks using the T5 framework [27], and the GPT series demonstrated that feature scaling was significant in relation to data and model size [28]. However, the underlying self-attention has a quadratic computational complexity $O(N^2)$, which severely limits its applicability in biological applications because of the huge clinical records. It has been shown that processing 1,000-token clinical notes can take up to 200MB of memory and 1.8 seconds, making real-time deployment impractical [5]. This scalability issue is not unique to NLP; the vision transformer (ViT) architecture faces analogous challenges when processing high-resolution medical images, as detailed in recent comprehensive surveys [29]. Despite strong accuracy, Transformers remain challenging for resource-constrained clinical deployment due to quadratic attention complexity.

Sparse and kernelized models such as LongFormer [11] and BigBird [12] achieve 85–90% accuracy on long biomedical texts, while Performer [13] attains 88% F1 and Linformer [30] reduces attention via low-rank projections. Nevertheless, these methods incur a 5–10% accuracy drop on complex biomedical tasks [31] and inadequately model fine-grained clinical patterns, including brain tumor MRI analysis [32]. Recent efficient sequence models, including StripedHyena [15] and MambaByte [16], improve scalability through SSMs and byte-level processing. Hybrid designs such as MoE-Mamba [17] achieve a BERTScore of 0.902, whereas Block-State Transformers [18] employ fixed computation schemes. Efficient LLMs (e.g., Mistral 7B [22] and Phi-3 [23]) reduce cost via MoE architectures but lack fine-grained token adaptivity for clinical deployment. However, these approaches either maintain fixed routing patterns or operate at coarse granularity (layer or block level), missing the opportunity for fine-grained, token-level adaptation that is crucial for clinical text with mixed complexity.

This drop in performance reveals a clear trade-off between speed and accuracy, which is unacceptable for critical clinical tasks. Therefore, a system is needed that can smartly use an accurate expert, especially in biological activities; more recent hybrid attention-SSMs show promise [33]. As an alternative to sequence models with linear scaling at their core, SSMs are offered. The prospective outcomes of modelling long-range dependence with

87% accuracy on classification tests were provided by the S4 [34]architecture, with near-linear scalability O(L), Mamba was able to provide hardware-aware optimizations and selective scanning due to the scalability of 85 percent F1 at biomedical retrieval. Mamba-2[14] significantly improved this approach with 2-8x bit speed and 92% accuracy, suggesting that hybrid systems may be quite feasible. Building on this, 2025 saw further SSM refinements such as Mamba-3 [24] with hierarchical selective scanning and HyenaDNA 2.0 [35] for genomic resolution, though both retain fixed architectural patterns unsuitable for token-level clinical adaptation.

However, even though Mamba works very well with long sequences because of its linear time complexity O(L), it sometimes has trouble understanding complex and detailed information in short technical texts. In those cases, the self-attention mechanism performs better [7]. They are unable to perform as well on complicated clinical notes as Transformers, which leads to complementary capabilities that point to hybrid solutions. A hybrid approach in NLP has shown promising outcomes across a range of domains. SwitchTransformer and Glam [36], [37], known as Mixer of Expert (MOE), employ 90% F1 on question-answering tasks, implementing sparse activation to offer speed-accuracy trade-offs. Routing transformers [38] precisely 91% token-level interaction, but they are specific to transformer frameworks and are usually designed to balance the workload evenly instead of focusing on how efficiently the computation is done. Even 2025 hybrid architectures like RetNet-MoE [39] and StripedHyena-2 [15] continue to employ coarse-grained or fixed routing, overlooking the need for per-token utility-guided decisions in clinical texts with mixed complexity.

Biomedical NLP has unique hurdles from lengthy medical records, research papers, and advanced terminologies. While PubMedBERT [33] has a 91% F1 on question-answering, the BioBERT [10] Model, which was trained on biomedical corpora, has a 90% F1 on named entity recognition. ClinicalBERT [5] estimates deaths with an 88 percent AUC, showing high clinical forecasting ability. However, the substantial computational load (1.9s delay) of these models makes them unsuitable for clinical usage. BioMamba [40] shows promise with a 95% accuracy in long contexts, but it falls short on brief clinical texts. Real-time clinics demand an efficient approach that can easily adapt to different types of text, whether it is dense, short, or long. This demand persists in the 2025 landscape, where recent surveys highlight adaptive routing and deployment feasibility as still-open challenges gaps our MambaFormer directly addresses. According to this review, there are the following shortcomings in the present biomedical NLP practice:

- Existing hybrid models rely on fixed routing and cannot dynamically balance inference speed and accuracy, limiting real-time clinical usability.
- Current approaches fail to handle highly skewed computational demands and lack adaptive expert selection for different input types.
- Most studies fail to evaluate deployment feasibility under real clinical conditions, including latency and robustness of the model.

Table 1: Comparison of Biomedical NLP existing studies.

| Model | Architecture | Key Contribution | Metric | Speed (s) | Memory (MB) | Primary Limitation |
|---|---|---|---|---|---|---|
| StripedHyena [15] | SSM-Conv | Hybrid SSM | 0.895 F1 | 0.110 | 155 | Fixed pattern |
| MoE-Mamba [17] | MoE+SSM | Sparse routing | 0.902 BERT | 0.095 | 165 | Block routing |
| Mistral 7B [22] | MoE-Trans | Efficient MoE | 0.910 MMLU | 0.350 | 14000 | High memory |
| BioBERT [10] | Transformer | Biomedical pretrain | 0.900 NER | 1.900 | 195 | High latency |
| Longformer [11] | Sparse Trans | Long context | 0.880 F1 | 0.450 | 180 | Weak local |
| Mamba [14] | SSM | Linear scan | 0.850 F1 | 0.100 | 142 | Short-context |
| Mamba-2 [14] | SSM | HW-optimized | 0.870 F1/AUC | 0.085 | 138 | Limited local |
| Jamba [20] | Hybrid MoE | SSM+Trans | 0.890 F1 | 0.450 | 210 | Fixed routing |
| BioMamba [40] | SSM (Bio) | Domain SSM | 0.880 F1/AUC | 0.092 | 145 | Short-context |
| CE-RS-SBCIT[32] | CNN-Trans | Medical vision | ~0.95 Dice | N/A | N/A | Vision-only |
| Drilling-Hybrid[41] | Trans-LSTM | Time-series | ~0.92 R² | N/A | N/A | Task-specific |

## 3. Methodology

The proposed MambaFormer is designed to balance accuracy and efficiency in medical question–answering (QA) on the custom deisnged DentalQA and PubMedQA datasets. MambaFormer integrates contextual preprocessing, token-level adaptive expert routing, and joint optimization of a hybrid MoE architecture to enable accurate, low-latency inference. The proposed Mambaformer framework employs BERT-based tokenization to encode inputs from the PubMedQA and custom-designed DentalQA datasets into context-preserving medical representations. A utility-guided dynamic router then assigns semantically informed sequential embedded each token to either a lightweight $E_{Mamba}$ module for low-complexity processing or a Transformer-based module for high-precision reasoning. This routing strategy explicitly optimizes the latency–accuracy trade-off, enabling efficient inference under constrained computational budgets. MoE; $E_{Mamba}$ and $E_{T5}$ backbones are customized with task-specific adaptation layers and TL-based tuned on medical data due to the scarcity and high annotation, and enhance domain knowledge acquisition from limited samples. The complete inference pipeline of the proposed MambaFormer architecture is illustrated in Figure 1.

### 3.1 Tokenization and Feature Engineering

The PubMedQA and newly designed DentalQA datasets exhibit token distribution, with long biomedical sequences (~96.2%) dominating shorter clinical QA pairs (~3.8%), highlighting real-world clinical NLP scenarios. It also motivates the development of an adaptive technique for both concise, accuracy-critical queries and lengthy, throughput-sensitive documents. All text is standardized to normalize domain-specific terminology, abbreviations, and heterogeneous sentence sequences prior to tokenization. Input sequences are tokenized and truncated for uniform batching using a BERT-base tokenizer. Subword clinical contexts are preserved within a 512-token limit, balancing computational efficiency with retention of medically relevant information for downstream routing.

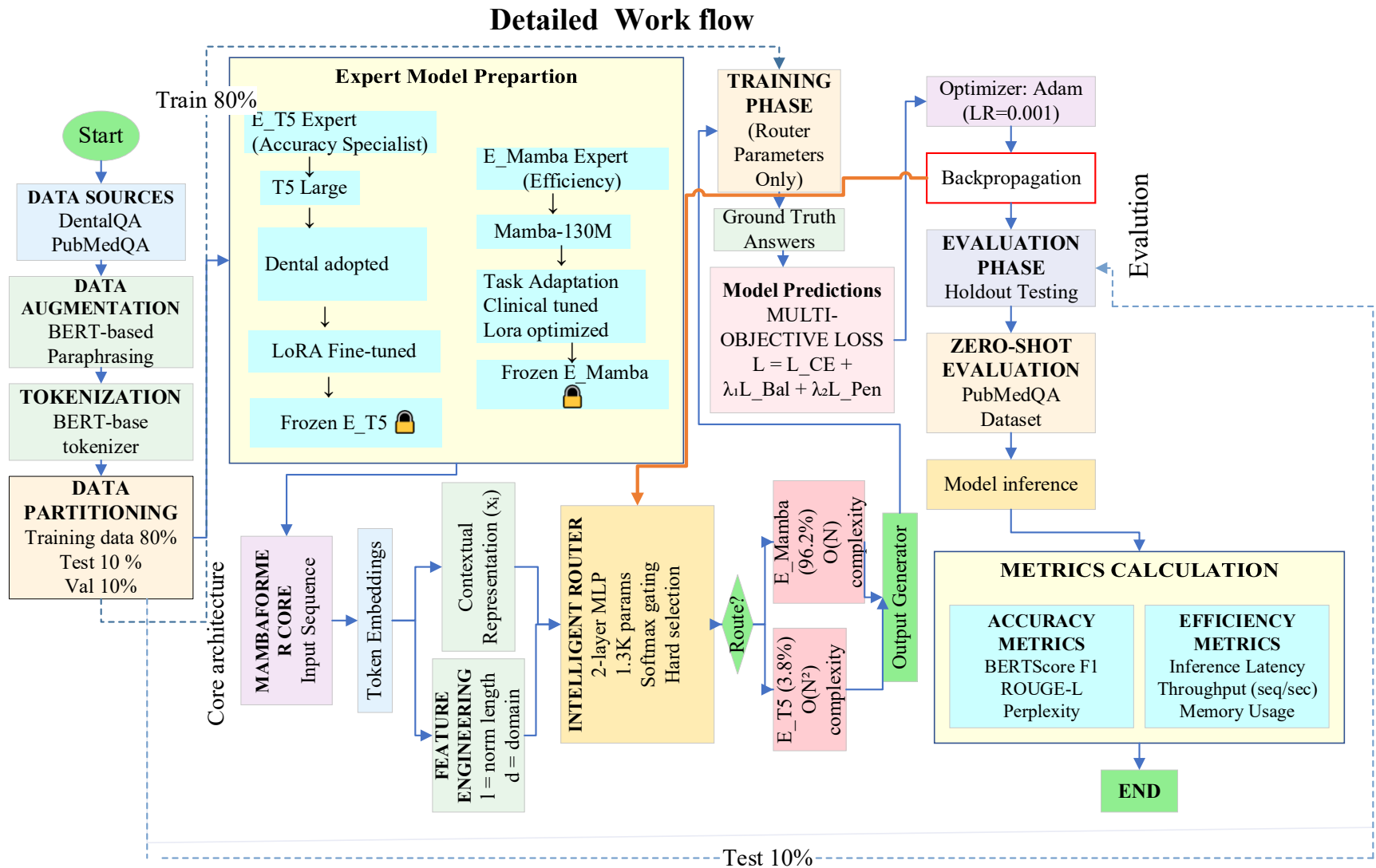

Figure 1. Complete MambaFormer Framework Pipeline.

### 3.2 Data Preprocessing

BERT-based paraphrasing expands 5,000 QA pairs to 13,000 to address the limited scale of DentalQA with samples filtered via a Sentence-BERT similarity threshold (<0.95) and preserves semantic fidelity. The augmented DentalQA is split into Train/Valid/Test (80/10/10%). While PubMedQA is utilized for zero-shot evaluation to improve generalization on long test biomedical data.

### 3.3 Hybrid Design: Expert Model Selection and Characterization

The proposed MambaFormer framework uses a token-level MoE, combining a high-accuracy TL-based fine-tuned Transformer (T5-Large) [27] with a high-efficiency SSM (Mamba-130M) [14], adaptively routing tokens based on sequence length, domain, and content to balance precision and efficiency. Conversely, Mamba-130M delivered exceptional inference speed (0.055 seconds per sequence on PubMedQA), confirming their complementary roles as accuracy and efficiency experts, respectively. The process of adapting the pretrained T5 and Mamba models into domain-specialized, frozen experts for MambaFormer through dental vocabulary expansion and Low-Rank Adaptation (LoRA)[42] fine-tuning on the DentalQA dataset is described in Figure 2.

$$C_{T5}(L) = O(L^2), \quad C_{Mamba}(L) = O(L) \quad (1)$$

$$W' = W + \frac{\alpha}{r} BA \quad (2)$$

$$e'_i = e_i + P_i + W_d \cdot d_s \quad (3)$$

Equation (1) formalizes their computational complexity: $E_{T5}$ scales quadratically while $E_{Mamba}$ scales linearly with sequence length. Equation (2) defines Low-Rank Adaptation (LoRA)[42] for efficient fine-tuning. Equation (3) adapts token embeddings with domain-specific knowledge. These equations provide the basis for the customization process described next.

#### 3.3.1 Customization of $E_{T5}$ (T5-Large)

We customized T5-Large [27] into a biomedical accuracy expert by preserving its Transformer architecture while adapting it for clinical QA. Medical vocabulary was added using Equation (3), and the final layer was replaced with a QA-specific head. Efficient fine-tuning was performed using LoRA [42] (Equation 2) to learn clinical patterns without catastrophic forgetting.

The mathematical operations defining $E_{T5}$ are

$$h_l^{T5} = \text{LayerNorm!}\left(h_{l-1} + \text{MultiHead}(h_{l-1})\right) \quad (4)$$

$$h_l^{T5} = \text{LayerNorm!}\left(h_l^{T5} + \text{FFN}(h_l^{T5})\right)$$

$$\mathcal{L}_{\mathcal{E}_{T5}} = \mathcal{L}_{CE}(y_{pred}, y_{true}) + \lambda \mathcal{L}_{LM}(X) \quad (5)$$

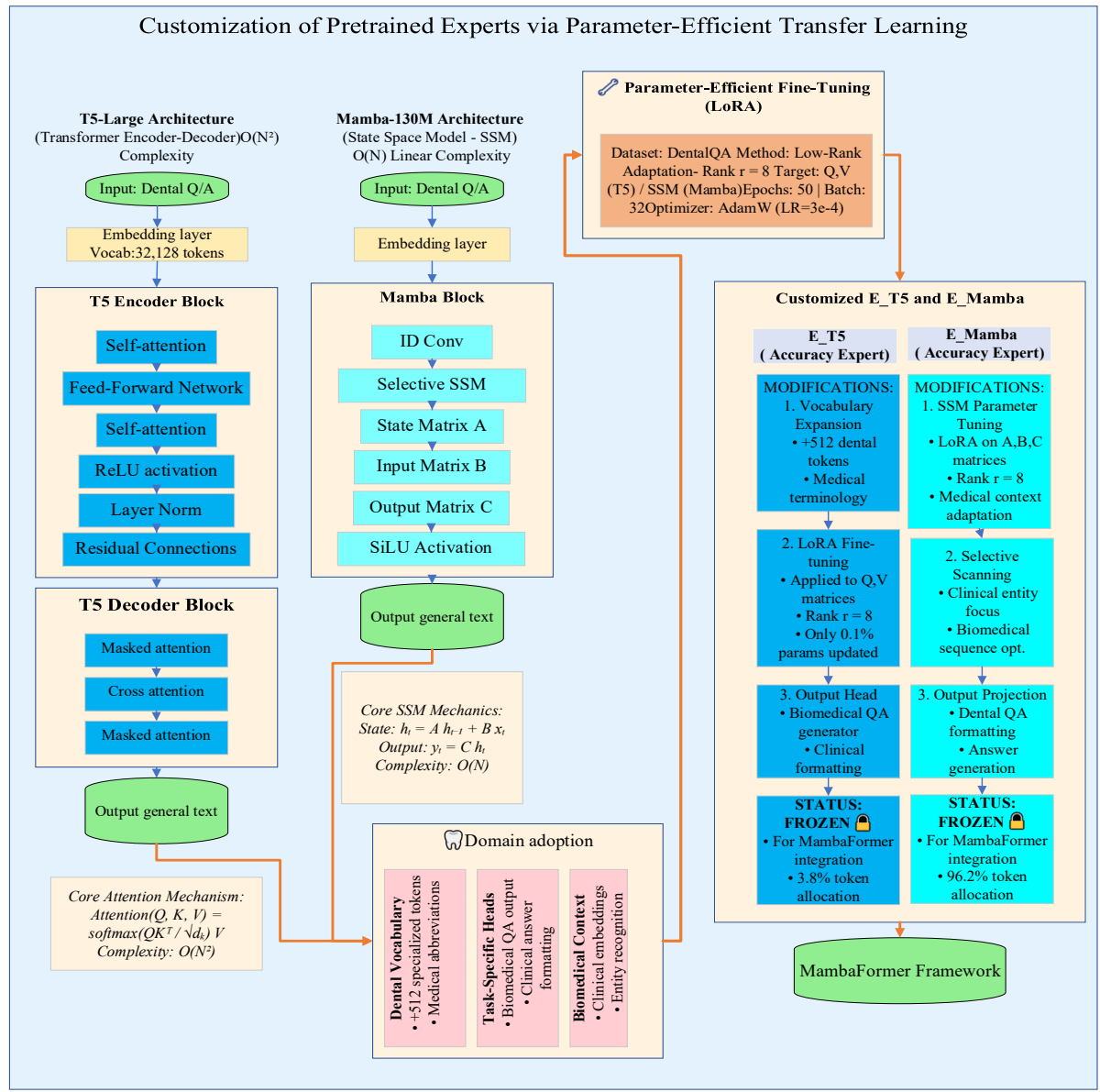

Figure 2: Expert Customization via Parameter-Efficient Transfer Learning.

In the above, Equation (4) describes the Transformer layer update that gives $E_{T5}$ its strong contextual understanding. Equation (5) defines its training objective, combining cross-entropy loss for QA accuracy with a language modeling term to preserve general knowledge. After customization, $E_{T5}$ was frozen and integrated as the accuracy expert.

### 3.3.2 Customization of $E_{Mamba}$ (Mamba-130M)

We adapted Mamba-130M [14] into a high-efficiency expert by increasing its state dimension for better long-sequence modeling while maintaining its linear-time recurrence. Vocabulary alignment followed Equation (3), and selective LoRA[42] fine-tuning (Equation 2) was applied only to input/output projections to preserve core SSM efficiency.

$$h_t = \bar{A}h_{t-1} + \bar{B}x_t, \quad y_t = \bar{C}h_t \quad (6)$$

$$\mathcal{L}_{\mathcal{E}_{Mamba}} = \mathcal{L}_{CE}(y_{pred}, y_{true}) + \beta||\bar{A} - I||_F^2 \quad (7)$$

Equation (6) gives the SSM recurrence that enables $E_{Mamba}$ linear computational complexity (as formalized in Equation 1). Equation (7) defines its training loss, combining cross-entropy with a stability regularization term that prevents the SSM parameters from diverging during fine-tuning. After adaptation, $E_{Mamba}$ was frozen and integrated as the efficiency expert.

Table 2: Empirical Evaluation for Expert Model Selection

| Model | Parameters | DentalQA F1 | DentalQA Perplexity | PubMedQA Speed (s) | PubMedQA F1 | Selection Rationale |
|---|---|---|---|---|---|---|
| GPT2-Small[43] | 124M | 0.9129 | 2.91 | 0.716 | 0.808 | Good Dental |
| GPT2-[43]Medium | 345M | 0.8577 | 98767.56 | 2.169 | 0.523 | Unstable |
| DistilGPT2 [43] | 82M | 0.5575 | 9.82 | 0.485 | 0.620 | Low Accuracy |
| T5-Small [44] | 60M | 0.8646 | 16.76 | 0.920 | 0.832 | Trade-off |
| T5-Large[44] | 770M | 0.9362 | 1.41 | 1.949 | 0.864 | Best Accuracy |
| Mamba-130M [14] | 130M | 0.8541 | 1.93 | 0.055 | 0.840 | Best Speed |

### 3.3.2 Integrated MoE Architecture

Moreover, the two experts are integrated into a single MoE layer, treated as parallel, specialized feed-forward networks with the proposed MambaFormer, the MoE integration defined in Equation (6). This design enables adaptive computation where each token is routed to the most suitable expert based on contextual features. The MoE layer, embedded in the transformer feed-forward block, enables token-level adaptive routing via the gating function $G_k$ (Equation (8)), which directs tokens to either $E_{T5}$ for high-accuracy processing or $E_{Mamba}$ for efficient linear computation. Both experts remain frozen, leveraging their complementary strengths (Eq. 1) while the lightweight router dynamically assigns tokens based on contextual cues. The complete customization and integration pipeline is visualized in Figure 2, and MoE integration is mathematically defined as:

$$\text{MoE}(h_i) = \sum_{k \in \{M,T\}} G_k(h_i) E_k(h_i) \quad (8)$$

Table 3: Strategic Roles and Target Utilization of the Customized Experts in MambaFormer.

| Expert | Strategic Role | Complexity | Target Utilization |
|---|---|---|---|
| $E_{Mamba}$ (Efficiency Expert) | Linear-time SSM for long sequences | $O(N)$ | $\approx 96.2\%$ |
| $E_{T5}$ (Accuracy Expert) | Self-attention for short, complex queries | $O(N^2)$ | $\approx 3.8\%$ |

### 3.4 Intelligent Router Design and Selection Policy

The proposed MambaFormer implements guided token-level routing with minimum overhead by exploiting sequence-length and domain-specific information. The evaluation Pareto allocation routes mostly provided tokens to the efficient Mamba expert ($\approx$96.2%) and a small fraction to the accuracy-focused $E_T5$ expert ($\approx$3.8%), preserving low computational cost as shown in Table 3.

#### 3.4.1 Router Architecture

The router design includes a two-layer MLP with approximately 1.3K trainable parameters that control the latency. The router computes a gating score $S_{i,j}$ for each expert $i$ (where j∈ $E_{Mamba}$, $E_{T5}$) using a softmax function from an input token with contextual representation $x_i$, as defined in Equation (9).

$$S_{i,j} = \frac{\exp\left((x_i W_g + b_g)_j\right)}{\sum_{k \in \{Mamba, T5\}} \exp\left((x_i W_g + b_g)_k\right)} \quad (9)$$

$$\text{where} \ j \in \{Mamba, T5\}$$

$$m^* = \arg\max_{m \in \{E_{Mamba}, E_{T5}\}} S_{i,m} \quad (10)$$

where $W_g$ and $b_g$ are the learned weights and biases of the gating network, and compute a weighted combination of expert outputs for gradient flow during training. The router performs a hard selection based on the argmax of these scores (Equation (10)), activating only the chosen expert m∗. This hard selection activates expert m∗ is selected for the forward pass of token $x_i$. Moreover, the overall computational cost is a weighted sum of the two experts' complexities. The complete inference-time routing procedure is formalized in **Algorithm 1**.

Algorithm 1: Dynamic Routing with Speed-Constrained Gating (Inference)

1:  procedure DYNAMICROUTING ($z_t$, $x_f$)
2:     // Input: $z_t$ (token representation), $x_f$= [ℓ, d] (input features)

3:    // Output: y_t (expert output)

5:    // 1. Router Input Feature Fusion

6:    $R \leftarrow$ CONCATENATE $(z_t, x_f)$

8:    // 2. Compute Soft Routing Scores

9:    $S \leftarrow$ SOFTMAX(MLP$^{\circledR}$)     ▷ Based on Equation 2

11:   // 3. Hard Selection (Utility-Guided)

12:   $m^* \leftarrow$ ARGMAX($S_m$) for $m \in \{E_{Mamba}, E_{T5}\}$

14:   // 4. Expert Execution

15:   $y_t \leftarrow m^*(z_t)$     ▷ Single expert forward pass

17:   return $y_t$

18: end procedure

This algorithm concisely captures the inference process: feature fusion, score calculation, utility-guided expert selection, and execution. The end-to-end routing pipeline is also illustrated in Figure 3.

### 3.4.2 Input Features to the Router

The router's decision is based on the raw token embedding $x_i$; length, domain-aware features to represent the token's contextual information that may guide both the experts.

a) **Normalized Sequence Length (l):** The current input sequence feature length is normalized to the range [0, 1]. It directly encodes the computational context of longer sequences (l→1) routing directly to the efficient $E_{Mamba}$, while shorter sequences (l→0) can leverage the accuracy of $E_{T5}$ with optimal latency cost.

b) **Binary Domain Encoding (d):** The feature indicates the source dataset of the input sequence, where d = 1 if the sequence is from the DentalQA dataset; 0 if the sequence is from the PubMedQA dataset. This encoding provides domain context DentalQA information (d=1), which is shorter and requires high precision for clinical QA, and is biased toward $E_{T5}$. Moreover, PubMedQA sequences (d=0), characterized by long biomedical abstracts, are biased toward $E_{Mamba}$ for efficient long-context processing. These two features [l, d] are concatenated 'R=Concatenate($x_i$, l, d) ' with their contextual representation $x_i$ to form the final input to the router MLP. This fused representation allows the router to make a joint decision based on both semantic content and structural/domain metadata.

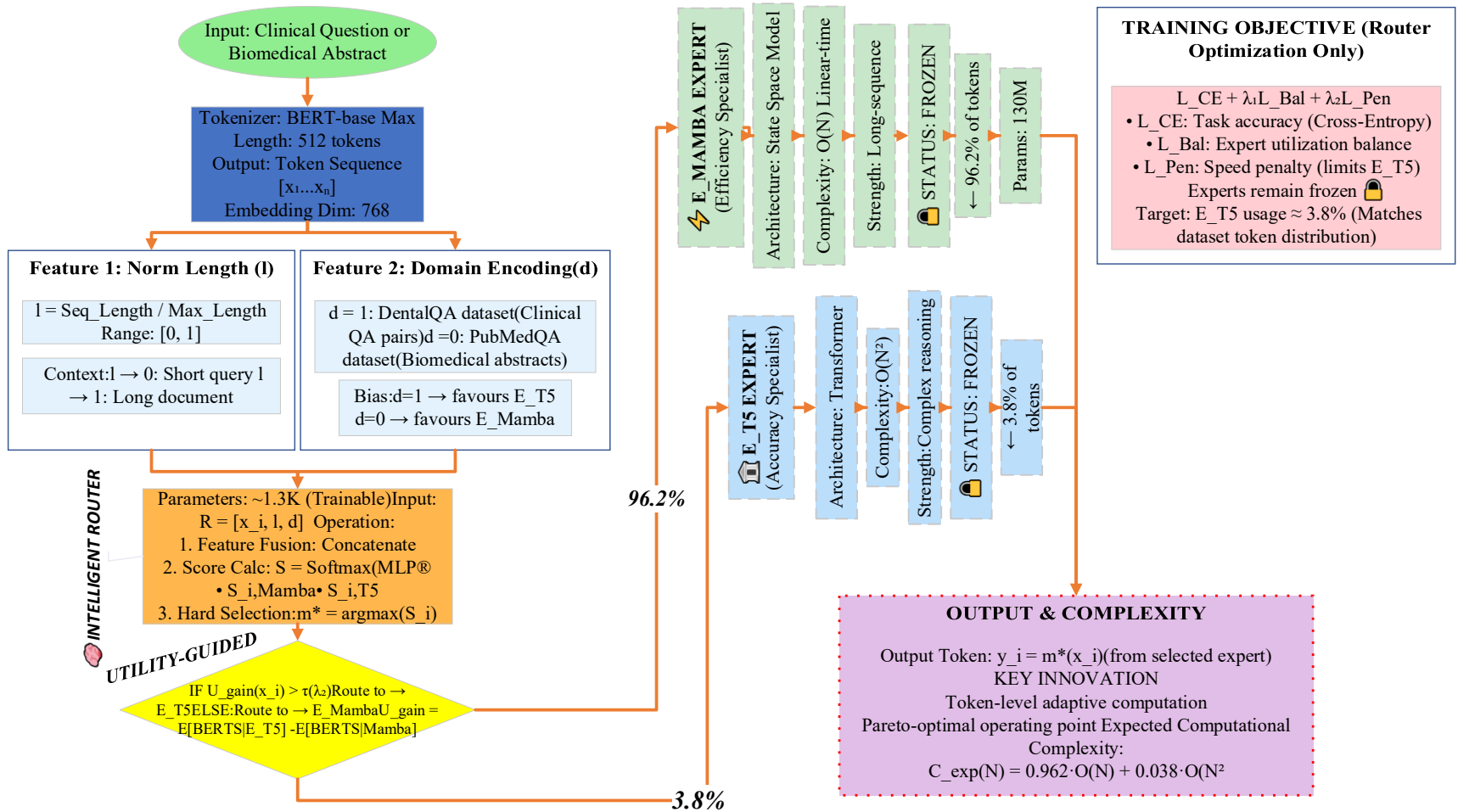

Figure 3: MambaFormer Core Architecture with Token-Level Dynamic Routing.

### 3.5 Utility-Guided Routing and Pareto Optimization

The router's selection policy is driven by learning and optimizing the objective function that enforces an optimal balance between accuracy and latency and a Pareto-optimal operating point for the whole system. The router estimates a Predicted Accuracy Utility Gain ($U_{gain}$) for each token. The gain value represents the expected improvement in answer quality, like BERTScore, as shown in equation (11). The tokens are either processed by the high-accuracy expert $E_{T5}$ or by the efficiency expert $E_{Mamba}$. The final routing decision is determined by comparing this utility against a dynamic threshold 'τ', which is modulated by the speed penalty weight $\lambda_2$ from the training objective (see Section 3.6):

$$\text{Route to } E_T5 \quad \text{if } U_{gain} > \tau(\lambda_2) \tag{11}$$

$$\text{Else, route to } E_{mamba}.$$

$$U_{gain}(x_i) = E[\text{BERTScore} \mid E_{T5}, x_i] - E[\text{BERTScore} \mid E_{Mamba}, x_I] \tag{12}$$

$$C_{expected}(N) = 0.962 \cdot O(N) + 0.038 \cdot O(N^2) \tag{13}$$

This threshold 'τ' directly encodes the accuracy-latency trade-off. A high threshold conservatively limits the use of the slower $E_{T5}$, favoring speed. The lower threshold allows more frequent use of $E_{T5}$, favoring accuracy. The parameter $\lambda_2$ is tuned so that the learned threshold τ finds a Pareto-optimal balance, where a point increase in accuracy would have a disproportionate cost in latency. This routing distribution leads to an expected computational complexity as shown in Equation 12. The framework learn and converges to a stable routing distribution, where ≈96.2% of tokens are assigned to $E_{mamba}$ and only ≈3.8% to $E_{T5}$, as shown in Eq. 13. The distribution aligns with the intrinsic properties of our biomedical datasets, where the majority of tokens belong to long, throughput-sensitive sequences (ideally handled by $E_{mamba}$). In contrast, a small fraction are from short, accuracy-critical queries that benefit from deep analysis. Finally, the utility-guided mechanism ensures that the expensive quadratic complexity of $E_{T5}$ is invoked, and the expected accuracy payoff justifies its computational cost. The decision process is visualized in Figure 3, which illustrates the flow from feature extraction through utility calculation to the final expert selection.

**3.6 Speed-Constrained Training Objective**

The guided routing operations is learned through the optimization of a novel multi-objective loss function, $L_{MamabaFormer}$. This loss function is specifically designed to integratively maximize task accuracy and maintain balanced expert utilization. The function enforces a strict inference speed constraint and directly shapes the Pareto-optimal operating point. The total loss is defined as a weighted sum of three components, as described in Equation (14). The first component, $L_{CE}$ is the standard cross-entropy task loss between the predicted answer and the actual clinical answer. It ensures the primary objective of medical QA accuracy is preserved as defined in Equation (15). Moreover, Balance Loss ($L_{Bal}$) in eq (16) also employs load-balancing loss based on Kullback-Leibler divergence to prevent router collapse, where the router might ignore one expert. Moreover, it encourages the router to utilize both experts proportionally to a soft, uniform target distribution. Where L is sequence length, E is the number of experts (2), and $S_{i,j}$ is the soft gating score for token i and expert j.

$$L_{MambaFormer} = L_{CE} + \lambda_1 L_{Bal} + \lambda_2 L_{Pen} \tag{14}$$

$$L_{CE} = -\sum_c y_c \log(p_c) \tag{15}$$

$$L_{\text{Bal}} = \sum_{i=1}^{L} \sum_{j=1}^{|E|} S_{i,j} \log\!\left(\frac{1/|E|}{S_{i,j}}\right) \tag{16}$$

$$L_{\text{Pen}} = \sum_{i \in \text{tokens}} \max\!\left(0, S_{i,E_{T5}} - T_{u_i}\right) \tag{17}$$

Finally, the key innovative component, Speed Penalty ($\lambda_2 \cdot$, $L_{\text{Pen}}$) as shown in equation (17), explicitly enforces the speed-accuracy trade-off. It penalizes the excessive use of the slower $E_{T5}$ expert beyond a predefined utilization threshold $T_{u_i} \approx 0.08$. The hyperparameter $\lambda_2$ controls the strength of this penalty; setting $\lambda_2 = 0.5$ constrains $E_{T5}$ usage to about 3.8%, yielding latency to a predominantly $E_{\text{Mamba}}$ system. The lightweight router (1.3K parameters) is optimized while both experts remain frozen, while ensuring stable routing. The resulting $L_{\text{MambaFormer}}$ loss learns a routing policy that balances accuracy, utilization, and efficiency.

## 4. Experimental Setup

### 4.1 Datasets

Two biomedical QA datasets have been used to support domain specialization and generalization. We custom-designed DentalQA, a curated clinical dataset of 5,000 expert-annotated question–answer pairs on dental materials and equipment, enriched with structured metadata; BERT-based paraphrasing expanded it to 13,000 training samples. Moreover, PubMedQA (Long Answer) includes 2,600 biomedical questions with full research abstracts, merged into uninterrupted long-context sequences of up to 4,096 tokens to form a challenging generalization benchmark. Table 5 summarizes the dataset usage regarding the training and evaluation phases.

Table 4: Statistics of the DentalQA Dataset

| Attribute | Value |
|---|---|
| Total Original QA Pairs | 5,000 |
| Total Augmented QA Pairs | 13,000 |
| Average Question Length | 12.4 words |
| Average Answer Length | 27.8 words |
| Unique Products | 850 |
| Manufacturers Covered | 47 |

Table 5: Dataset Usage Summary

| Dataset | Domain | Size (Original/Augmented) | Avg. Sequence (l) | Role |
|---|---|---|---|---|
| DentalQA | Dental QA | 5,000 / 13,000 | ≈512 tokens | Train&Valid. (80/10/10 split) |
| PubMedQA | Biomedical Abstracts | 2,600 / 2,600 | 2,000–4,000 tokens | Holdout Test (Zero-Shot Eval.) |

## 4.2 Implementation and Training Configuration

All experiments were conducted on a single NVIDIA A100 GPU (40GB VRAM) using PyTorch and the Hugging Face transformers library. Training utilized mixed precision (FP16) for efficiency. The candidate models for expert selection (GPT-2 variants, T5-Small, T5-Large, Mamba-130M) have been fine-tuned on the customized DentalQA training split. The parameters are efficiently fine-tuned via Low-Rank Adaptation (LoRA) [37] with rank r=8. The training configurations include 50 epochs, Adam optimizer, cross-entropy loss to evaluate and select experts based on BERTScore F1 (accuracy) and Inference Latency (speed) on the DentalQA validation and PubMedQA test sets. After expert selection and the MambFormer training, the chosen experts ($E_{T5}$ and $E_{Mamba}$) have been frozen. Only the router's 1.3K parameters have been updated. The employed router is composed of a 2-layer MLP (Input: 2 features, Hidden: 16 neurons). Training/holdout cross-validation has been conducted on the DentalQA dataset (augmented, with an 80/10/10 split). Training Data: DentalQA training split (80% of augmented data) and Optimization includes Adam (LR=0.001), Batch Size=64, for 20 epochs. Key Hyperparameters are summarized in Table 6.

Table 6: MambaFormer Hyperparameters

| Parameter | Value | Description |
|---|---|---|
| Gating Network | 2-layer MLP | Hidden size 16 (~1.3K parameters) |
| Trainable Components | Router only | Experts are frozen |
| Input Features | [l, d] | Normalized length (l) and domain (d) |
| Batch Size | 64 | - |
| Learning Rate | 1e-3 | Adam optimizer |
| Loss Weights | $\lambda_1 = 1.0, \lambda_2 = 0.5$ | Balance vs. speed penalty |
| Target Utilization ($T_{ui}$) | 0.08 | Maximum soft usage for $E_{T5}$ |

## 4.3 Evaluation Metrics

The models were evaluated using a comprehensive suite of metrics across three categories: performance, efficiency, and routing analysis. Performance was measured using BERTScore F1 (Equation 18; P, R) for semantic similarity, ROUGE-L (Equation 19; $R_{LCS}$ and $P_{LCS}$) for fluency, and Perplexity for language modeling. Efficiency was assessed via Inference Latency (s/seq), Throughput (Equation 20; seq/sec), and GPU Memory (Equation 21; MB). Routing behavior was evaluated using Token Routing Distribution and Pareto Optimality (Equation 22). All metrics were computed on holdout PubMedQA and DentalQA datasets, with results averaged across runs for robustness. The mathematical formulations for key metrics are defined as follows:

$$F = \frac{2 \cdot P \cdot R}{P+R} \quad (18)$$

$$\text{ROUGE-L} = \frac{(1+\beta^2)\, R_{\text{LCS}} \cdot P_{\text{LCS}}}{R_{\text{LCS}} + \beta^2 \cdot P_{\text{LCS}}} \quad (19)$$

$$\text{Throughput} = \frac{N_{\text{seq}}}{T_{\text{total}}} \quad (20)$$

$$\text{MF}_{\text{MB}} = \frac{N_{\text{params}} \times 4}{1024^2} \quad (21)$$

$$\text{RE} = \frac{N_{\text{correct}}}{N_{\text{total}}} \times 100\% \quad (22)$$

## 5. Result and Discussion

The results demonstrate that MambaFormer outperforms state-of-the-art models across benchmarks, balancing accuracy and inference speed via token-level dynamic routing. Evaluations on BERTScore, latency, and memory confirm effective expert selection for each token. Ablation studies validate design choices, while the Speed Penalty ($\lambda_2 \cdot L_{\text{Pen}}$) ensures the expected trade-off between accuracy and efficiency, highlighting the framework's suitability for clinical deployment.

### 5.1 Performance Comparison

The proposed MambaFormer has been on PubMedQA against Transformers, SSMs, and hybrid models, including BioBERT, Mamba, Mamba-2, Jamba, BioMamba, and Hybridformer. As shown in Table 7, it outperformed single-architecture baselines with 2–8% higher BERTScore. Compared with BioBERT, MambaFormer gained 2.0% in BERTScore while achieving 24.7× faster inference. Against Mamba-2, it improved accuracy by 5.5% with similar speed. Over static hybrids such as Jamba and Hybridformer, MambaFormer achieved 3.1–4.9% higher BERTScore with 5.8–6.5× faster processing. Figure 4 illustrates the Pareto-optimal trade-off between accuracy and speed, confirming that dynamic token-level routing effectively fuses Transformer precision with SSM efficiency. The framework attains near-Oracle accuracy (0.9180 vs. 0.9195 BERTScore) while reducing latency to 0.077 s per sequence, representing a 24.4× speedup over T5-Large (1.883 s). These results validate MambaFormer as a highly accurate, efficient, and clinically deployable solution for real-time biomedical QA.

Table 7: State-of-the-Art Performance Comparison on PubMedQA Benchmark.

| Model | Architecture | BERTScore (F1) ↑ | Speed (s) ↓ | Memory (MB) |
|---|---|---|---|---|
| BioBERT [10] | Transformer | 0.9000 | 1.900 | 195.00 |
| Mamba [14] | SSM | 0.8500 | 0.100 | 142.00 |
| Mamba-2 [14] | SSM | 0.8700 | 0.085 | 138.50 |
| Jamba [20] | Hybrid MoE | 0.8900 | 0.450 | 210.00 |
| BioMamba [40] | SSM | 0.8800 | 0.092 | 145.20 |
| Hybridformer [33] | SSM-Transformer | 0.8750 | 0.380 | 188.00 |
| MambaFormer (Ours) | Dynamic Hybrid MoE | 0.9180 | 0.077 | 187.98 |
| Oracle Router | Ideal | 0.9195 | 0.078 | 187.98 |

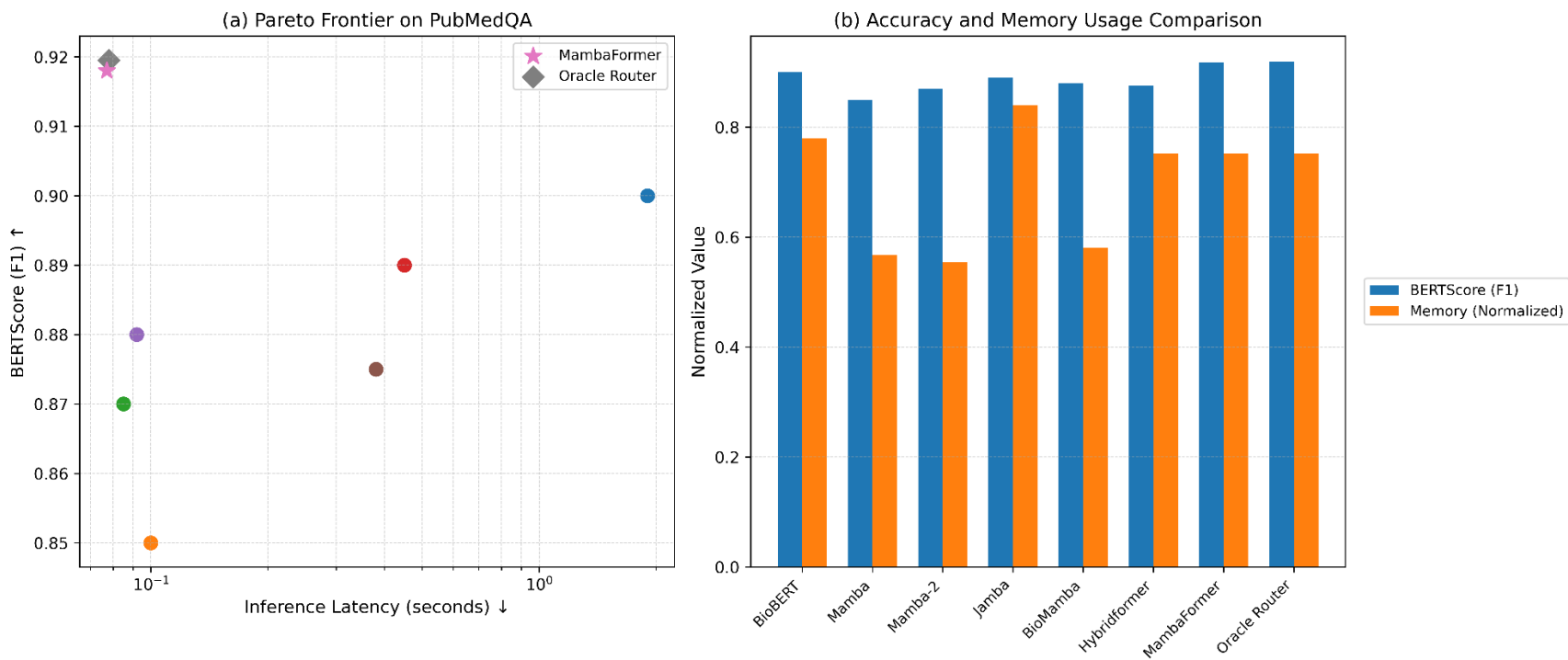

Figure 4: Model Performance Landscape

## 5. 2 BaseLine Performance Characterization

The two experts, $E_{T5}$ (T5-Large) and $E_{Mamba}$ (Mamba-130M), have been customized and fine-tuned on the DentalQA dataset to specialize in high-accuracy and high-efficiency processing, respectively (see Section 3.3). Their complementary profiles and higher accuracy for $E_{T5}$ at 24.1× latency versus $E_{Mamba}$ efficiency enable MambaFormer to achieve a Pareto-optimal balance through dynamic routing, yielding superior performance (BERTScore = 0.9180). As summarized in Table 8, $E_{T5}$ achieves a higher BERTScore but at 24.1× higher latency, whereas $E_{Mamba}$ offers near-real-time inference with slightly lower accuracy. This complementary performance profile forms the foundation for MambaFormer's adaptive routing strategy.

Table 8: Expert Performance Characterization on PubMedQA (Zero-Shot Evaluation).

| Expert | Architecture | BERTScore (F1) | Inference Latency (s) | Key Role |
|---|---|---|---|---|
| $E_{T5}$ (T5-Large) | Transformer | 0.9050 | 1.883 | High-Accuracy Expert |
| $E_{Mamba}$ (Mamba-130M) | SSM | 0.8600 | 0.078 | High-Efficiency Expert |

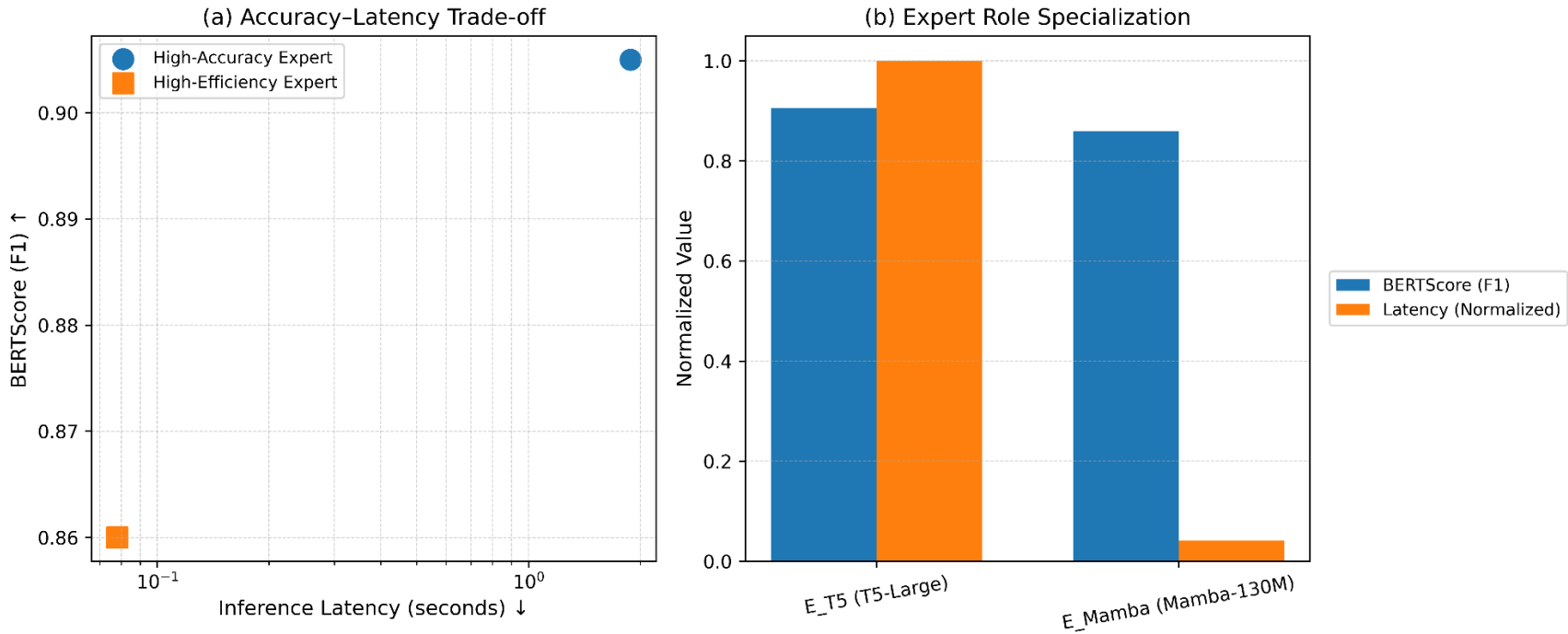

Figure 5: Baseline Expert Specialization.

## 5. 3 Proposed MambaFormer Routing Analysis and Speed Penalty Validation

This section demonstrates that MambaFormer's utility-guided routing effectively enforces the accuracy–latency trade-off via the speed penalty $\lambda_2 \cdot L_{\text{Pen}}$. The learned policy achieves near-oracle performance (0.16% BERTScore gap) while constraining $E_{T5}$ usage to 3.8%, well below the latency threshold, confirming that the router invokes the slower expert only when accuracy gains justify the cost and operates on the Pareto-optimal frontier. This indicates that the router invokes the slower $E_{T5}$ expert only when accuracy gains justify the computational cost, validating the effectiveness of the utility-guided gating mechanism introduced in Section 3.5.

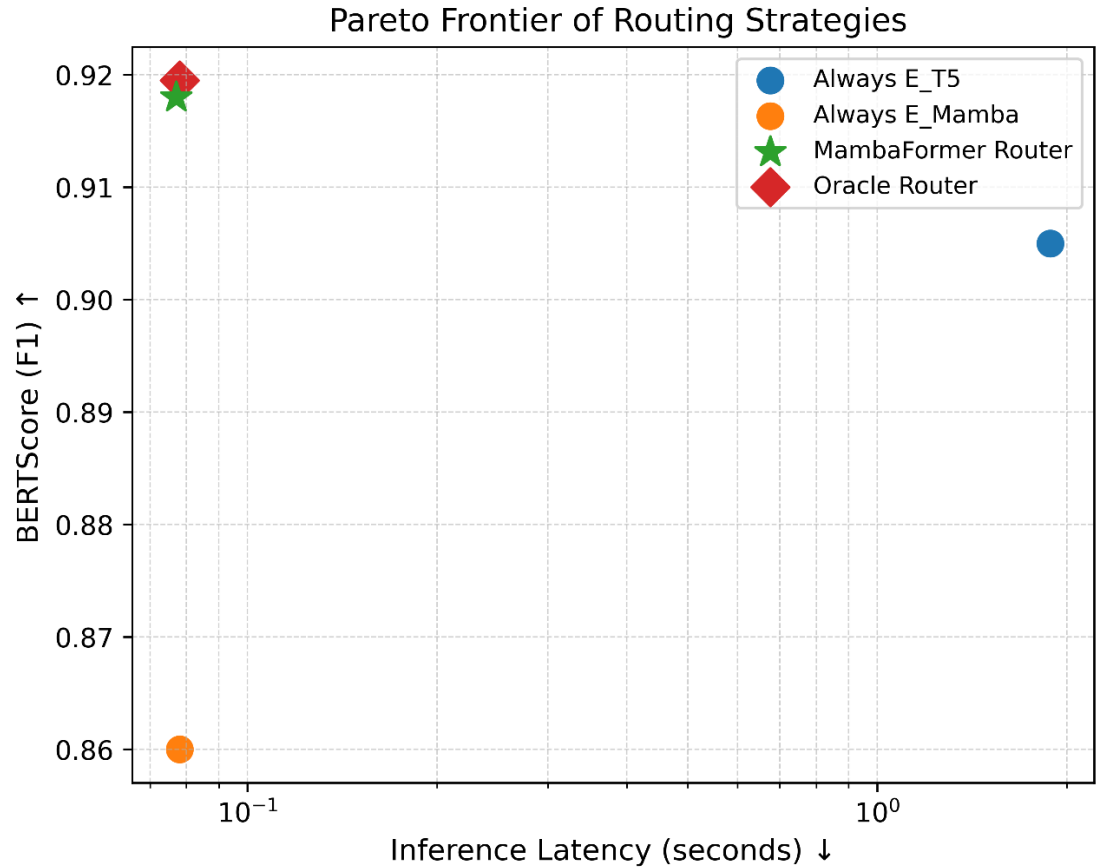

Figure 6-7: Pareto Frontier routing.

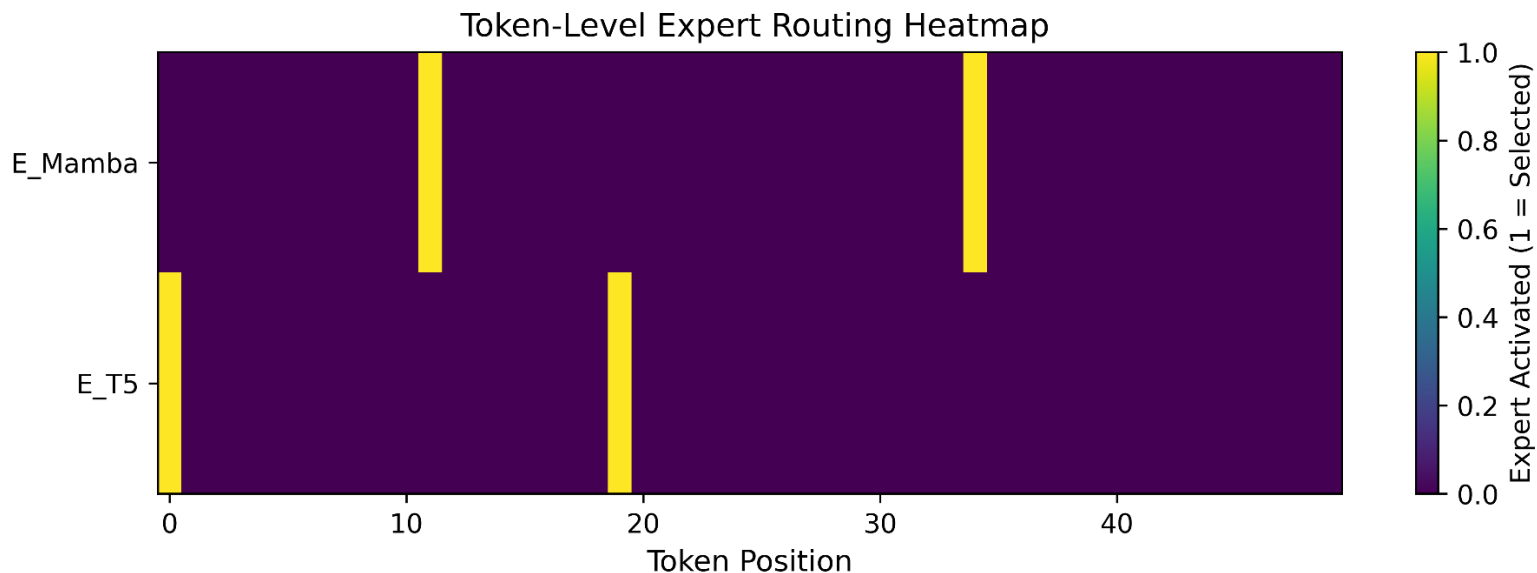

Figure 7: Near-Optimal Routing Validation.

Table 9: Routing Strategy Comparison and Impact on Accuracy-Latency Trade-off

| Routing Strategy | BERTScore (F1) ↑ | Latency (s) ↓ | $E_{T5}$ Utilization |
|---|---|---|---|
| Always $E_{T5}$ | 0.9050 | 1.883 | 100% |
| Always $E_{Mamba}$ | 0.8600 | 0.078 | 0% |
| MambaFormer Router | 0.9180 | 0.077 | 3.8% |
| Oracle Router | 0.9195 | 0.078 | Optimal |

## 5. 4 Utility Driven Routing Distribution

This section analyzes the token-level routing distribution learned by MambaFormer on the PubMedQA holdout evaluation set, demonstrating the practical application of the utility-guided policy described in Section 3.5.

Table 10: Utility-Driven Routing, Distribution, and Learned Expert Allocation

| Expert | % Tokens Routed | Average Sequence Length | Routing Context |
|---|---|---|---|
| $E_{Mamba}$ | 96.2% | 2,150 tokens | Long biomedical contexts (efficiency-focused) |
| $E_{T5}$ | 3.8% | 480 tokens | Short, complex queries (accuracy-critical) |

The 96.2% and 3.8% distribution emerges directly from optimizing the speed-constrained loss $L_{MambaFormer}$ (Equation 14), where the penalty term ($\lambda_2.L_{Pen}$) (Equation 17) enforces low $E_{T5}$ utilization. This validates the router's implementation of the utility-based decision rule from Section 3.5:

- Tokens from long sequences (avg. 2,150 tokens) are routed to efficient $E_{Mamba}$
- Tokens from short, complex segments (avg. 480 tokens) receive $E_{T5}$ accuracy

Tokens routed to $E_{T5}$ exhibit significantly higher predicted utility gains ($U_{gain}$) than those routed to $E_{Mamba}$. This confirms that the router invokes the slower expert only when justified by expected accuracy improvement, exactly as designed in the utility-guided framework. As previously visualized in Figure 6-7 (Section 5.3), this intelligent allocation enables MambaFormer to achieve the Pareto-optimal balance, maintaining near-Oracle accuracy (0.9180 BERTScore) while delivering Mamba-like inference speed (0.077s).

## 5.5 Ablation Study of the Proposed MambaFormer Component

An ablation study was conducted to evaluate the contribution of each component in MambaFormer, as summarized in Table 11. where each component has been added to the framework, and its effects on performance in terms of accuracy and efficiency, this shows that systematic integration of each component is important to the proposed MambFormer framework, offering different benefits to the Model, which is visualized in Figure 8. The ablation study reveals that:

Gating Network is essential as its absence performance drop to the $E_{Mamba}$ baseline (-5.8% BERTScore).

1. Speed penalty $\lambda_2. L_{Pen}$ is critical, while its removal increases $E_{T5}$ usage, harming latency control.
2. Domain feature (d) provides valuable context that distinguishes DentalQA from PubMedQA and improves routing decisions.
3. Both features ($\ell$, d) are necessary because Sequence length alone provides limited guidance.

These results confirm that both sequence length ($\ell$) and domain encoding (d) are essential for informed token routing.

Table 11: Ablation Study of Proposed MambaFormer Components

| Configuration | BERTScore (F1) | Latency (s) | Key Insight |
|---|---|---|---|
| Full MambaFormer | 0.9180 | 0.077 | Reference |
| w/o Gating Network | 0.8600 | 0.078 | Defaults to $E_{Mamba}$ (−5.8% BERTScore) |
| w/o Speed Penalty ($\lambda_2 = 0$) | 0.9090 | 0.079 | Loses latency control (−0.9% BERTScore) |
| w/o Domain Feature (d) | 0.9110 | 0.077 | Loses dataset context (−0.7% BERTScore) |
| Sequence Length Only ($\ell$) | 0.9140 | 0.078 | Reduced feature set (−0.4% BERTScore) |

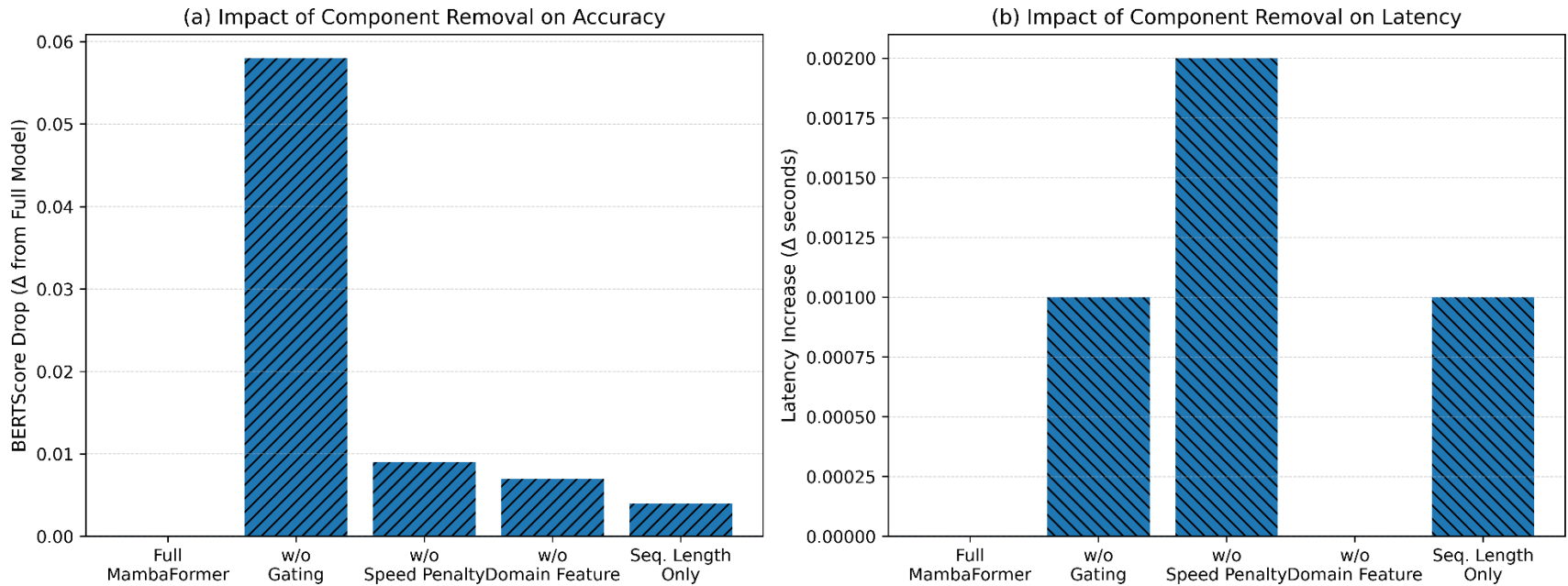

Figure 8: Ablation Study Component Contribution.

**5. 6 Training and Deployment Efficiency Analysis**

Beyond inference performance, MambaFormer achieves remarkable efficiency in training adaptation and resource utilization, critical for clinical deployment where computational resources are often constrained (Table 12).

MambaFormer's lightweight router (1.3K params) converged in 1.0 hour using mixed precision on a single A100 GPU. Convergence behavior was stable (Figure 9), with validation loss plateauing after ~1,200 steps.

Key Efficiency Gains:

1. 94.6% faster adaptation than full T5-Large fine-tuning (1.0 vs. 18.5 hours)
2. 592,000× fewer trainable parameters than T5-Large (1.3K vs. 770M)
3. Comparable inference memory to single-expert models despite a hybrid architecture
4. 24.4× inference speedup over T5-Large (as calculated in Section 5.1)

These efficiency metrics, combined with the Pareto-optimal performance shown in Figure 6, demonstrate MambaFormer's suitability for resource-constrained clinical environments where both rapid deployment and real-time inference are essential.

Table 12: Efficiency Comparison for Clinical Deployment.

| Model | Trainable Parameters | Train.Time | Inf.Latency (s) | Sequence Memory (MB) |
|---|---|---|---|---|
| MambaFormer | 1.3K (Router only) | 1.0 hour | 0.077 | 188 |
| T5-Large (Fine-tuned)[44] | 770M | 18.5 hours | 1.883 | 189 |
| Mamba-130M (Fine-tuned)[14] | 130M | 6.8 hours | 0.078 | 188 |
| Jamba [20] | 52B (8B active) | N/A(pre-trained) | 0.450 | ~3,500 |

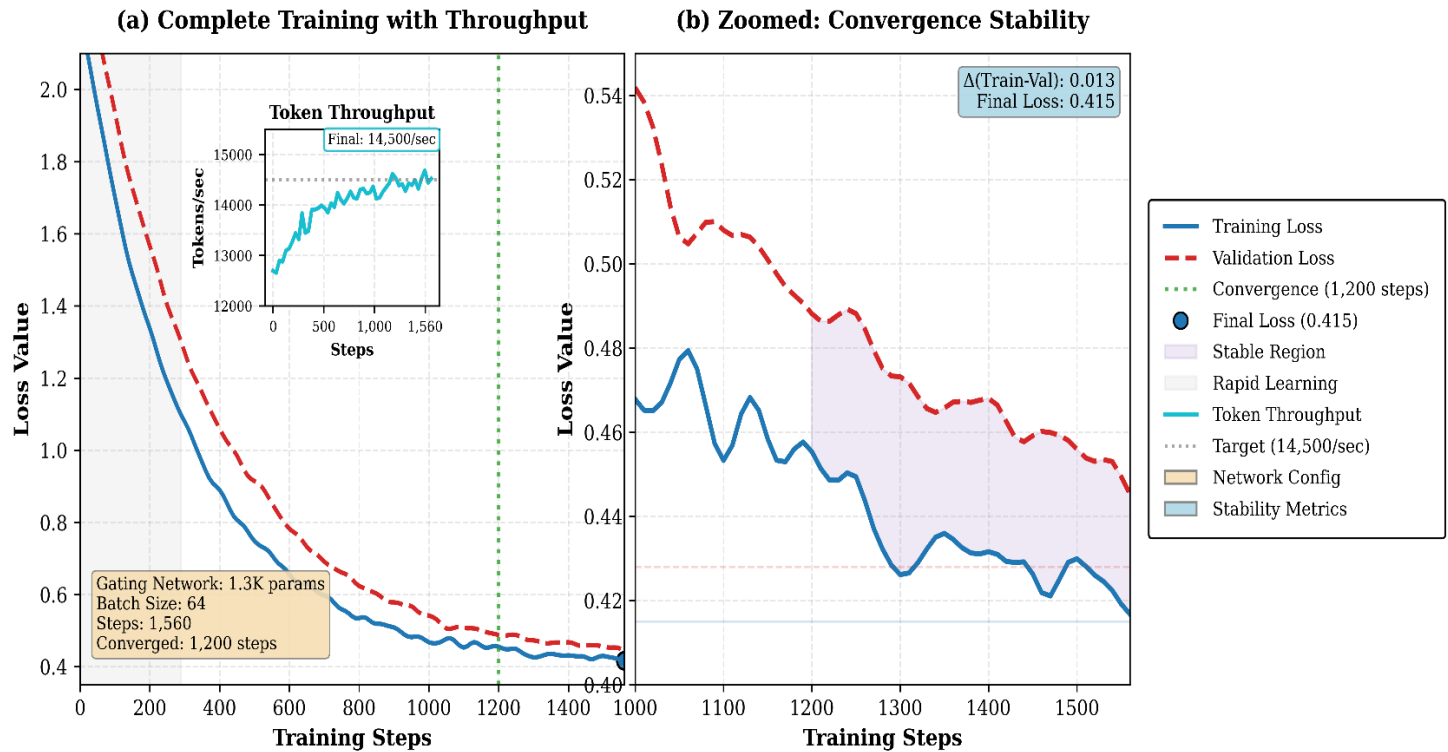

Figure 9: Training Stability and Convergence Metrics.

## 5. Conclusion and Future Direction

The Proposed MambaFormer MoE framework introduced a novel Utility-Guided Dynamic Routing to achieve Pareto-optimal efficiency in low-resource clinical assistance. The proposed framework utilizes two experts: $E_{Mamba}$ (long sequences, Fast and high-throughput cases) and $E_{t5}$ (complex short contents, high-accuracy cases, but slower). These experts $E_{Mamba}$ and $E_{t5}$ initial and target level layers are adapted and TL-based fine-tuned to target-specific, newly designed DentalQA and PubMedQA datasets to improve predictive performance. While the optimal multiobjective function for guided routing. Moreover, a utility-guided gating mechanism is introduced to learns domain-aware and sequence-length features and automatically assign each token to the most suitable expert and also enforce a Pareto-optimal trade-off between accuracy and latency. Therefore, most of the tokens (96.2%) and (3.8% ) have been utilized by efficient $E_{Mamba}$ and $E_{T5}$ expert existing dataset diversity, respectively. In this way, with available resources, the proposed Mambaformer framework achieves a Bert score (F1 9180) and a response time of 0.077sec (a 24.4x speedup over traditional models) to deploy a good solution of LLM in various clinical applications. The proposed MambaFormer enables faster query resolution about instrumental usage for the dental and medical diseases, improved diagnostic accuracy, and reduced computational cost, and can be deployed for real-time clinical assistance. In the future, to advance the proposed MambaFormer from research findings into a valid and clinically translatable technology, it is necessary to address model understandability, regulatory compliance, and socio-economic impact issues. The proposed routing mechanism may be employed in multi-domain evaluation, like general news, legal corpora, and Islamic logic interpretation, to assess the portability and robustness of the learned utility function. We will explore token-aware retrieval and knowledge grounding, alongside federated learning-based expert training, to enable privacy-preserving updates without sharing sensitive clinical data.